\documentclass{article}

\usepackage[main, final]{neurips_2026}

\usepackage[utf8]{inputenc} 
\usepackage[T1]{fontenc}    
\usepackage{url}            
\usepackage{booktabs}       
\usepackage{amsfonts}       
\usepackage{nicefrac}       
\usepackage{microtype}      
\usepackage{xcolor}         

\usepackage{graphicx}
\usepackage{amsmath}
\usepackage{amssymb}
\usepackage{amsthm}
\usepackage[most]{tcolorbox}
\usepackage{mathtools}
\usepackage{wrapfig}
\usepackage{pifont}
\usepackage{caption}
\usepackage{multirow}
\usepackage{wrapfig}
\usepackage[dvipsnames,svgnames]{xcolor}
\definecolor{lightyellow}{RGB}{255,244,204}
\usepackage{colortbl}
\definecolor{citecolor}{HTML}{2980b9}
\definecolor{linkcolor}{HTML}{c0392b}
\usepackage[colorlinks,citecolor=citecolor,linkcolor=linkcolor]{hyperref}

\RequirePackage{xspace}
\makeatletter
\DeclareRobustCommand\onedot{\futurelet\@let@token\@onedot}
\def\@onedot{\ifx\@let@token.\else.\null\fi\xspace}
\def\eg{\emph{e.g}\onedot}
\def\ie{\emph{i.e}\onedot}
\def\modelname{MARS}

\newtheorem{theorem}{Theorem}[section]

\newtheorem{lemma}[theorem]{Lemma}

\title{Does Seeing More Mean Knowing More? Mono-Anchored Advantage Normalization for Multi-Source Visual Reasoning}

\author{
Fanhu Zeng\textsuperscript{\rm 1}~~~
Zhicong Luo\textsuperscript{\rm 2}~~~
Zefan Wang\textsuperscript{\rm 1}~~~
You Li\textsuperscript{\rm 3}~~~
Chi Chen\textsuperscript{\rm 1}\thanks{Correspondence to: Chi Chen <chenchithu@gmail.com>}~~~ 
Maosong Sun\textsuperscript{\rm 1}~~~\\
  \textsuperscript{\rm 1}Tsinghua University~~~~~~
  \textsuperscript{\rm 2}Northwestern Polytechnical University \\
  \textsuperscript{\rm 3}Beijing Jiaotong University\\
}

\begin{document}
\maketitle

\begin{abstract}
Visual reasoning through reinforcement learning with verifiable rewards~(RLVR) has achieved remarkable progress. However, when dealing with multi-source inputs, existing approaches tend to treat them as a mere accumulation of information, lacking explicit mechanisms to distinguish whether integrating additional sources yields information gain or introduces interference.
Therefore, they struggle to effectively model dynamic interaction when integrating multiple sources, particularly when they differ significantly in physical properties and semantics, \eg, infrared and depth, leading to inferior performance to mono-source reasoning when a certain source holds the dominant signal.
To address this issue, we propose \modelname{}, a novel mono-anchored multi-source reasoning framework that models each visual modality as an independent information source. Specifically, by treating mono-source rewards as dynamic anchors, our method explicitly incorporates the information gain introduced by multi-source fusion into advantage normalization and adaptively emphasizes mutual promotion between sources while suppressing potential noise or conflicts during RLVR.
From theoretical analysis, our method effectively quantifies information gain introduced by multi-source integration in gradient estimation, enabling consistent modality regulation. Empirical results also show impressive 3.2\% and 4.9\% performance gains on GRPO and DAPO across diverse datasets, confirming the effectiveness of our method. Code is available at \url{https://github.com/AI9Stars/MARS}.
\end{abstract}

\section{Introduction}
Recent advances in multimodal large language models~(MLLMs), which align representations across vision and language modalities~\cite{bai2025qwen3}, have demonstrated strong capabilities in multimodal perception and understanding~\cite{li2025migician}. 
More recently, visual reasoning~\cite{li2026imagination, xu2025llavacot} has been introduced to encourage deeper thinking through reinforcement learning with verifiable rewards~(RLVR), allowing models to generate structured responses with self-reflection through explicit reasoning rather than direct prediction, thereby fostering the emergence of chain-of-thought~(CoT) reasoning~\cite{wei2022chain} and enhancing the ability of complex understanding, multi-step reasoning, and logical consistency.

\begin{wrapfigure}{r}{0.6\linewidth}
    \centering
    \vspace{-8pt}
    \includegraphics[width=1.0\linewidth]{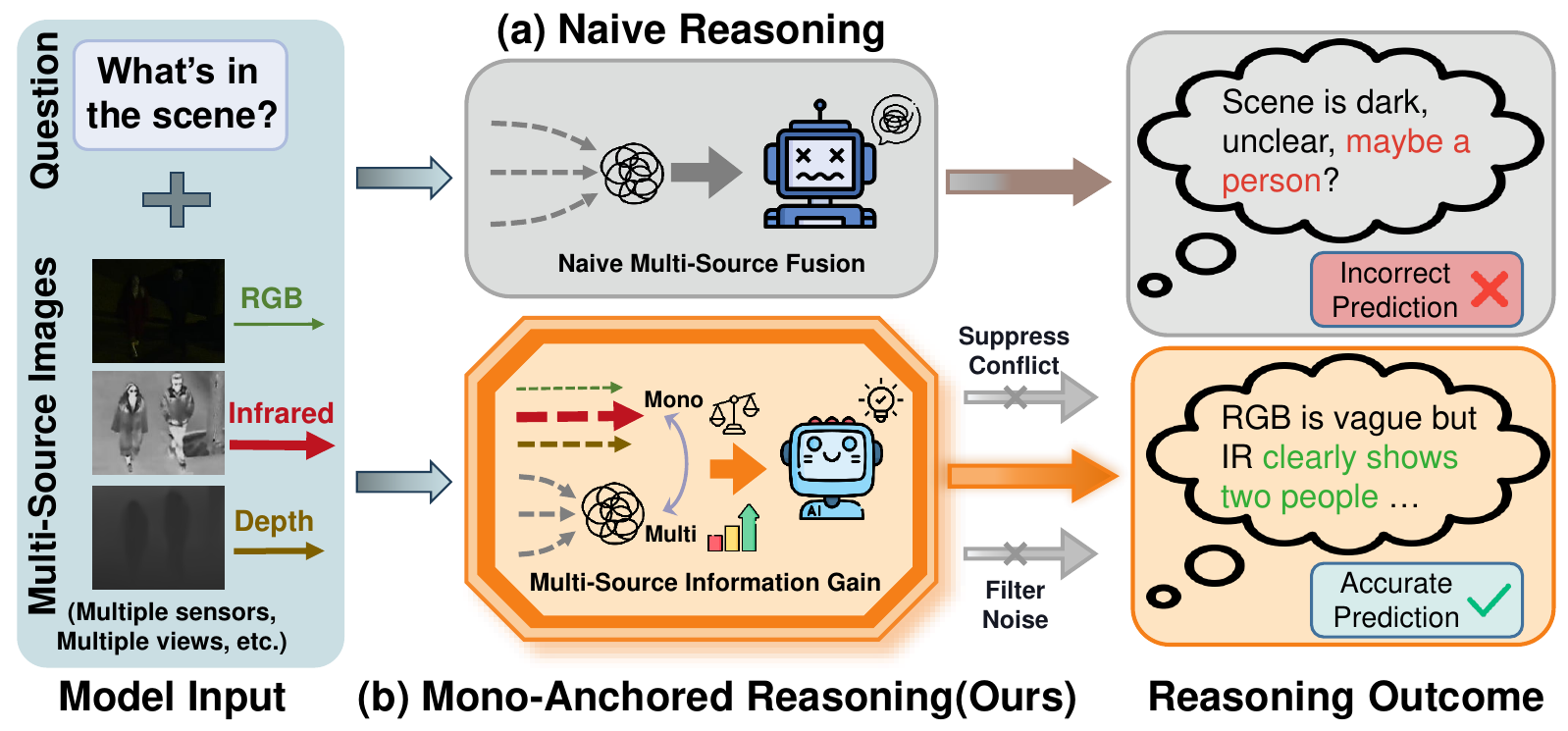}
    \caption{Illustration of multi-source visual reasoning. (a) Existing methods struggle to model dynamic interaction in multi-source data; (b) our method explicitly uses mono-source rewards as anchors to measure the information gain from multi-source integration and enhances reasoning.}
    \vspace{-8pt}
    \label{fig:illustration}
\end{wrapfigure}

Despite the progress of visual reasoning, current methods largely optimize for aligned representations, and the complementary strengths of different sources are often assumed and overutilized, \ie, seeing more modalities means knowing more, but potential interference or conflicts are seldom explicitly explored.
In particular, existing RLVR frameworks optimize multi-source rewards directly, without explicitly assessing whether integrating additional sources yields positive information gain or instead introduces interference relative to strong mono-source reasoning, especially when their attributes and semantics have significant differences, such as medical imaging~\cite{azam2022review}, autonomous driving~\cite{caesar2020nuscenes}, remote sensing~\cite{zhang2010multi}, and so on. 
In these scenarios, \textit{\textbf{naively integrating multiple sources can even lead to performance inferior to strong mono-source reasoning}}, when a specific source contains the dominant and reliable signal. 
Therefore, there is a lack of principled mechanisms to model dynamic interaction, whether complementary or conflicting, between multiple sources, which may become particularly evident in complex scenarios where a critical cue resides in a specific source.
As shown in Figure~\ref{fig:illustration}, in tasks where inherent physical limitations and degradation are caused by illumination variation, occlusion, and adverse weather conditions, relying solely on RGB imagery or the relationship between sources is often inadequate. In contrast, different sources such as infrared, depth, or multi-view can provide crucial and robust information with more reliable scene understanding, which requires handling multi-source data in a comprehensive manner.

In this paper, we aim to enhance the ability of visual reasoning when dealing with multi-source data. Based on the analysis, we uncover that a core reason for this limitation lies in the naive way current visual reasoning frameworks handle source integration. Specifically, they treat all inputs equally contributive to the task, and fail to explicitly model the performance interactions between specific source and multi-source data, which is a key factor for effective multi-source regulation.
From an optimization perspective, these interactions correspond to whether multi-source reasoning improves or degrades performance relative to mono-source baselines, a distinction that remains invisible to advantage estimation in existing RLVR frameworks.
This gap motivates the need for a general approach that can dynamically regulate contributions from certain sources.

To this end, we propose \modelname{}, a novel multi-source reasoning framework that explicitly incorporates each visual modality as an individual information source and models the information gain introduced by multi-source integration.
Concretely, by treating mono-source rewards as the anchors, it computes advantages based on information gains between multi-source and mono-source\footnote{They specifically describe features of different visual modalities that differ in physical
properties and semantics.} rewards. We theoretically analyze that our method guarantees and enables dynamic optimization that emphasizes promotion while suppressing noisy or conflicting information during training by maximizing multi-source information gain. Notably, our algorithm enhances multi-source utilization from inherent capability without architectural redesign, offering a general and effective solution for improving visual reasoning performance.

We conduct experiments on various multi-source tasks, including depth, infrared, multi-view and text-rich understanding. Extensive results with notable \textbf{3.2\%} and \textbf{4.9\%} improvements on GRPO and DAPO and in-depth analyses strongly validate the effectiveness and generalizability of our method. Our contributions are summarized as follows:

\begin{itemize}
    \item We reveal that existing multi-source visual reasoning can systematically degrade performance, and identify relative information gain over mono-source reasoning as the key factor for effective multi-source integration from theoretical derivation.
    \item We design a novel visual reasoning method that introduces mono-source rewards as anchors to quantitatively measure multi-source information gain from integration in advantage normalization, enabling adaptive regulation of different sources during RLVR training.
    \item We conduct extensive experiments on various multi-source visual reasoning tasks, and the consistent and significant performance improvements on different RL algorithms validate the effectiveness and generality of our approach.
\end{itemize}

\section{Related Work}
\noindent \textbf{Reinforcement Learning with Verifiable Rewards}
has made substantial progress in recent years, with pioneering systems such as DeepSeek-R1~\cite{guo2025deepseek} and Kimi~\cite{team2025kimi} demonstrating that complex reasoning patterns can emerge through optimization with verifiable rewards, where outcome reward signals are used to guide the learning of long reasoning chains. Within this paradigm, some approaches focus on enhanced optimization strategies~\cite{zheng2025group,zhang2025r1}, such as regularization, stabilized policy updates, and refined reward designs, to improve consistency and robustness.
Building on these foundations, visual reasoning incorporates images into reasoning by coordinating linguistic reasoning with perceptual states. It achieves strong performance in vision-centric tasks such as grounding~\cite{bai2025univg} and image understanding~\cite{yang2025r1}, highlighting it as a promising paradigm for complex multimodal understanding and deduction. In this paper, we focus on the capability of visual reasoning with multi-source data.

\noindent \textbf{Multi-Source Visual Reasoning} refers to tasks that require a joint understanding of images from multiple sources, potentially captured from different sensors, times or viewpoints~\cite{zhang2018multi}. This is crucial for real-world intelligent systems, where a single source is often insufficient for achieving completeness and reliable decisions in complex environments. Early studies focus on multi-source fusion~\cite{brenner2023rgb, yuan2024improving}, where extracted features from different cameras are explicitly fused to enhance robustness and geometric consistency. More recently, multimodal large language models have reframed multi-image reasoning as unified and aligned comprehension with implicit correspondences and shared representations. Nevertheless, current attention is paid to general domain enhancement and evaluation~\cite{fu2024blink, yu2024spark}, which overlooks the complementarity and contradictions of multi-source data in the reasoning process.
In practical applications, it is crucial for models to determine the reliability of multiple visual inputs, rather than relying on manually selecting the most salient modality, which is valuable in critical fields such as medical diagnosis~\cite{huang2026echomllm} and robotic manipulation~\cite{ni2026semanticvla}.

\section{Methodology}
\subsection{Motivation}
\label{sec:motivation}
Visual reasoning has exhibited strong understanding capabilities under multi-image inputs. However, we observe a consistent and non-trivial phenomenon: as depicted in Figure~\ref{fig:illustration}, when handling images from multiple sources, \eg, infrared, depth and so on, and only one image among multiple sources is truly informative for the task, typical multi-source reasoning often fails to capture and concentrate on the critical visual scene and therefore underperforms the upper bound of mono-source reasoning, even if all available sources are provided jointly.
This contradicts the cognition of humans that integrating more information always brings more knowledge, and naturally raises an open question: 

\textit{Does seeing more mean knowing more in multi-source visual reasoning? If not, how can we solve it?}

We attribute the issue to potential modality interference in multi-source reasoning. Specifically, a typical visual reasoning model is normally trained under the assumption of complementary data integration, \ie, seeing more images brings more knowledge, and only learns the positive guidance of multi-image fusion with implicitly unified and shared representations.
Without explicitly identifying which image is causally responsible for correct decisions, it therefore struggles to capture the dynamic interaction, \ie, promotion or interference between modalities. 
This is especially severe in multi-source scenarios where images have different properties and semantics, resulting in unstable or noisy learning dynamics.
Therefore, advantage estimation becomes unreliable under such conflict, where standard advantage normalization estimates statistics solely from multi-source trajectories, and may be dominated by spurious correlations introduced by non-informative sources.

At this point, specific mono-source reasoning often provides a significant and stable inductive signal in these scenarios. When the key image is present, a specific mono-source rollout tends to produce more consistent reward with semantic information, effectively guiding the optimization direction. 

To this end, we propose to incorporate mono-source rollouts into the advantage estimation of multi-source rollouts. Intuitively, mono-source reasoning acts as a general dynamic anchor to stabilize and guide multi-source reinforcement learning: (1) if it underperforms with modality conflicts, the algorithm softly regularizes trajectory updates toward the more reliable mono-source behavior; (2) moreover, if multi-source reasoning outperforms mono-source reasoning with modality mutual promotion, the algorithm also encourages exploration beyond mono-source cues.

\begin{figure}[t]
    \centering
    \includegraphics[width=0.88\linewidth]{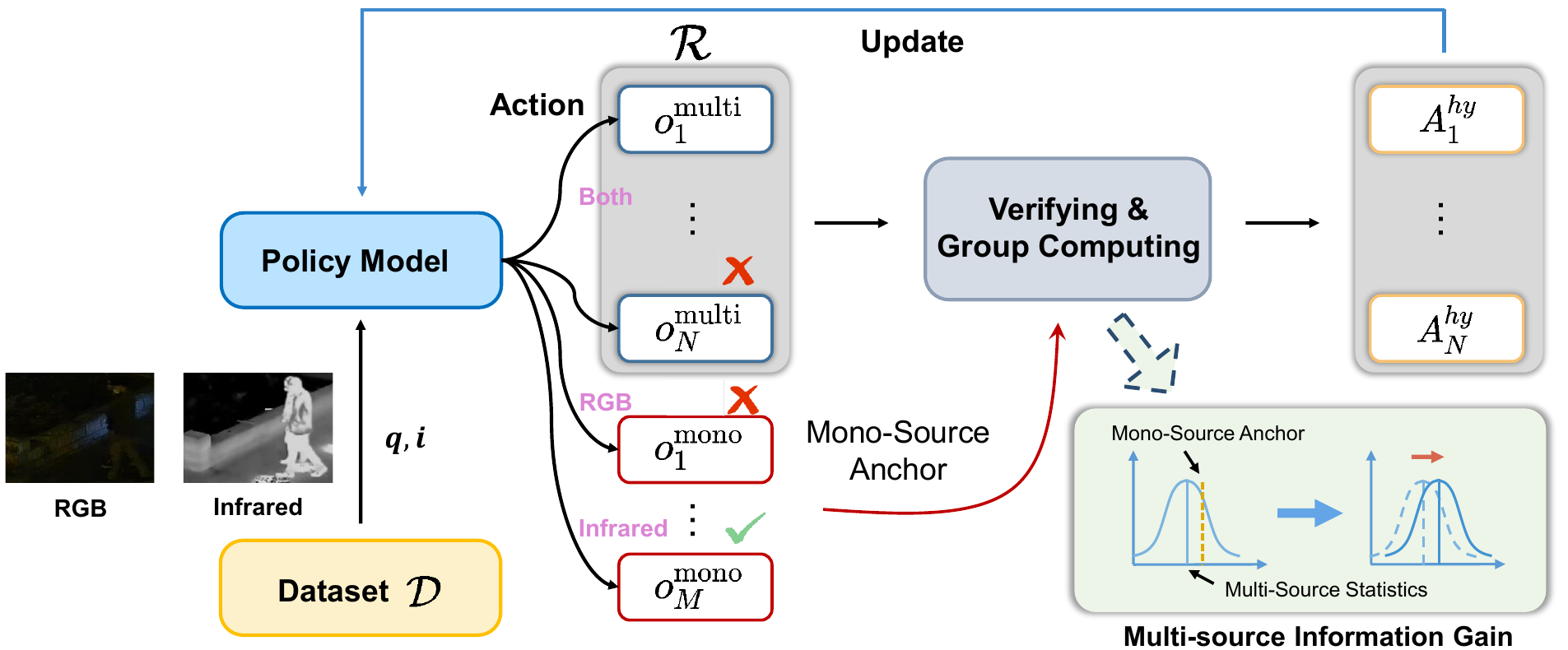}
    \caption{Structure of the proposed mono-anchored advantage normalization for multi-source visual reasoning. Mono-source rewards serve as dynamic anchors to quantify the influence of source integration with multi-source information gain in on-policy optimization.}
    \label{fig:structure}
    \vspace{-14pt}
\end{figure}

\subsection{\modelname{}: Mono-Anchored Advantage Normalization for Multi-Source Reasoning}
\label{sec:method}
\noindent \textbf{Preliminary.}
For each instance consisting of question $q$ and multiple images $i$ in training dataset $\mathcal{D}$, multi-source trajectories~(rollouts) are generated through policy $\pi_\theta$ parameterized with $\theta$, where multiple images are jointly provided as input:
\begin{equation}
\mathcal{G}^{\text{multi}}=\left\{ o^{\text{multi}}_{j} \right\}_{j=1}^{N}.
\end{equation}
The reward $r$ is exploited to measure the output in response to input and each rollout is normalized by group-wise mean and variance of reward to obtain advantage for stability:
\begin{equation}
    A_j= \frac{r(q,i,o_j^{\text{multi}})-\mu^{\text{multi}}}{\sigma^{\text{multi}}}.
\end{equation}

The standard policy gradient algorithm optimizes the expected advantage function $J(\theta)$, and its policy gradient estimator~\cite{sutton1998reinforcement} has the following form:
\begin{equation}
    \nabla_\theta J(\theta) = \mathbb{E}_{\{q,i\}\sim  \mathcal{D}, \ o \sim \pi_\theta(q,i) } [A \cdot \nabla_\theta \log \pi_\theta(o|q,i)],
\end{equation}
where $\{q,i\}$ is the question and image from dataset $\mathcal{D}$, and the policy $\pi_\theta$ generate the trajectories for verifiable reward.

\noindent \textbf{Advantage Normalization with Mono-Source Anchor.}
As illustrated in Figure~\ref{fig:structure}, in terms of reasoning with multi-source visual tasks, motivated by the function of mono-source rewards in advantage estimation, we additionally generate mono-source rollouts, where each image is individually paired with the textual input to produce rewards with the same policy model:
\begin{equation}
\mathcal{G}^{\text{mono}}=\left\{ o_j^{\text{mono}} \right\}_{j=1}^{M}.
\end{equation}
In terms of advantage estimation, it is performed for multi-source rollouts only, while leveraging hybrid statistics of both multi and mono-source rewards for gradient estimation to stabilize the normalization:
\begin{equation}
    A^\text{hy}_j= \frac{r(q,i,o_j^{\text{multi}})-\mu^\text{hy}}{\sigma^\text{hy}}, j=1,\cdots,N.
\end{equation}
Specifically, mono-source rollouts are not used to directly update the multi-source policy. Instead, their role is to adjust the normalization statistics as an adaptive reference. Intuitively, as illustrated in the right of Figure~\ref{fig:structure}, when multi-source reward outperforms mono-source reward, the introduced estimation would lower the mean for multi-source enhancement. Conversely, if a particular modality plays a decisive role, our algorithm will inhibit the model from learning contradictory multi-source rewards and instead drive it toward better modality-specific learning.

\noindent \textbf{Verifiable Rewards.} The verifiable reward is a key component in reinforcement learning to align the preferences of models, which may include simple verification functions~\cite{shao2024deepseekmath} that check whether predictions match the correct answers in contents and formats. Applying this concept to visual tasks requires adaptation of specific rule-based verifiable reward functions. \textbf{For grounding tasks}, grounding reward is directly formulated by calculating the average Intersection-over-Union~(IoU) between predicted and ground truth bounding boxes:
\begin{equation}
    r_\text{iou}(q,i,o) = \frac{1}{K}\sum_{i=1}^{K}IoU_i,\\
\end{equation}
where $K$ is the number of objects in the scene, and the grounding reward consists of the iou reward and the format reward:
\begin{equation}
    r_{\text{grounding}} = r_{\text{iou}} + r_{\text{format}}.
\end{equation}

\begin{figure}[t]
    \centering
    \includegraphics[width=1.0\linewidth]{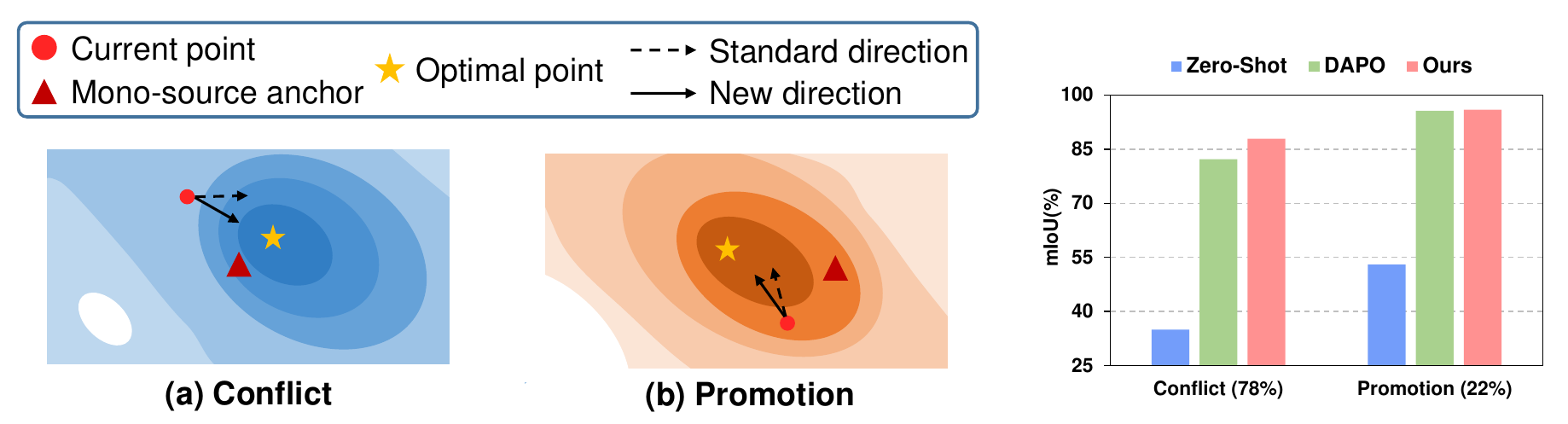}
    \vspace{-12pt}
    \caption{Illustration of mono-source anchor in optimization. It provides a consistent facilitation in both modality conflict and promotion. This results in significant gain across scenarios. }
    \label{fig:optimization-illusrtation}
    \vspace{-10pt}
\end{figure}

\textbf{For visual question answering tasks}, the accuracy reward is determined by whether the output matches the ground truth:
\begin{align}
r_{\text{acc}}(q,i,o) = \left\{
\begin{array}{ll}
1 & , \text{if} \quad \text{ground truth in} \ o,\\
0 & , \text{otherwise}. \\
\end{array}
\right.
\end{align}
The final reward is a combination of accuracy reward and format reward:
\begin{equation}
    r_{\text{vqa}} = r_{\text{acc}} + r_{\text{format}}.
\end{equation}

\noindent \textbf{Remark.} In the case of multi-source visual reasoning, instead of estimating the baseline from multi-source rewards alone, our algorithm computes hybrid statistics over the union of mono-source and multi-source rewards:
\begin{equation}
    \mathcal{R}
    =
    \left\{
    o^{\text{mono}}_1, \dots, o^{\text{mono}}_M,
    o^{\text{multi}}_1, \dots, o^{\text{multi}}_N
    \right\},
\end{equation}
from the same policy model. This can be seen as on-policy optimization with a hybrid distribution. Concretely, by leveraging mono-source rewards as anchors, our method precisely utilizes the difference between trajectories, thereby enhancing performance with exact information gain from mono-source to multi-source rewards shown in Figure~\ref{fig:step}\textcolor{linkcolor}{b}.

\subsection{Theoretical Analysis}
\label{sec:theory}
We present several theoretical analyses of the proposed mono-anchored advantage normalization algorithm from policy optimization perspective to construct key properties for stability and rationality.
\begin{theorem}[Unbiasedness]
\label{the:unbiasedness}
    For any measurement of gradient estimation, the expectation of our algorithm for is equivalent to the expectation of on-policy optimization:
\begin{equation}
    \mathbb{E}_{q,i\sim  \mathcal{D}, \ o \sim \pi_\theta(q,i) }[A^\text{hy}\cdot \nabla_\theta\log \pi_\theta(\cdot|q,i)] = \nabla_\theta J(\theta).
\end{equation}
\end{theorem}
This provides a theoretical guarantee that the proposed algorithm introduces no bias for gradient estimation under the condition of on-policy optimization from the perspective of expectation, ensuring stability. Appendix~\ref{app:proof1} provides the detailed proof.

\begin{theorem}[Gradient Decomposition]
\label{the:decom}
    The gradient optimization based on mono-anchored advantage normalization is equivalent to maximizing the multi-source information gain while optimizing the standard multi-source reward:
\begin{equation}
    \nabla_\theta J^\text{hy}(\theta) = \nabla_\theta J(\theta)+(1-\alpha)\Delta_{IG}\nabla_\theta J^{\text{reg}}(\theta),
\end{equation}
\end{theorem}
where conventional advantages have a zero mean. $\alpha=N/(M+N)$ is the proportion of multi-source rollouts representing the strength of guidance. $\Delta_{IG}=\mu^{\text{multi}}-\mu^\text{mono}$ is the expectation of the reward increment of multi-source trajectories compared to mono-source trajectories. It measures the relative information gain from multiple image fusion relative to mono-source reasoning, and a negative value indicates conflict between modalities that perform inferior to mono-source reasoning. 

The derivation holds for any $\alpha \in[0,1]$ and justifies the practical effectiveness of the advantage estimation scheme combining both sample types through unified standardization. This reveals that the practical effect of \modelname{} arises from the finite-sample gradient estimator used by RL algorithm that our algorithm is theoretically optimizing a weighted multi-source reward with a multi-source information gain regularization, which emphasizes trajectories consistent with reliable sources, while suppressing those affected by cross-source interference, thereby reweighting the sampled gradients. Specifically, by leveraging the mono-source reward as an explicit and systematic signal, it dynamically adjusts the standard gradient direction according to multi-source information gain during the optimization procedure, which guides the optimization towards an optimal point with faster convergence and better multi-source performance. Figure~\ref{fig:optimization-illusrtation} gives an intuitive illustration of mono-source anchor in optimization. In conflict where certain source performs well, it is around the optimal point, and pull the optimization direction close by information gain. This greatly improves the performance, where standard multi-source reasoning undergoes a severe performance drop. The process is similar in promotion that pushes the direction away from mono-source anchor towards the optimal point, providing a consistent facilitation in multi-source visual reasoning.
The quantitative results show the commonalities of conflict and the utility of our method in resolving conflict rather than a general effect. The proof is shown in Appendix~\ref{app:proof2}.

\begin{figure}[t]
    \centering
    \includegraphics[width=0.98\linewidth]{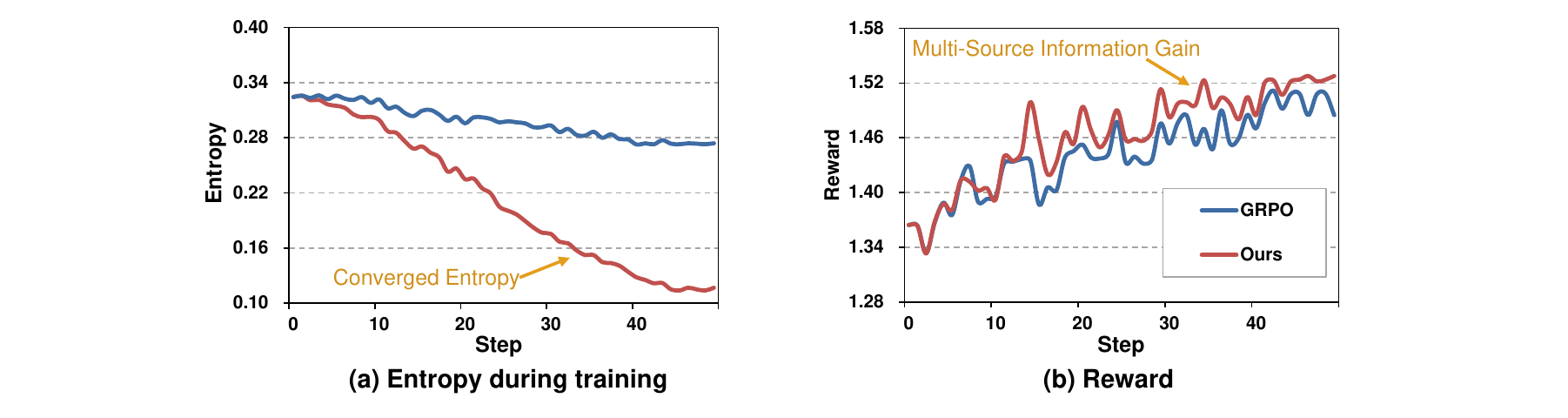}
    \vspace{-10pt}
    \caption{Learning statistics of (a) entropy and (b) reward when training vanilla GRPO and our algorithm. Our method achieves better performance with multi-source information gain from a mono-source anchor while maintaining stability.}
    \vspace{-15pt}
    \label{fig:step}
\end{figure}

\noindent \textbf{Remark.} Inherently, we study policy optimization with a mono-anchored advantage for multi-image reasoning. Instead of universally increasing rewards or advantages, it enforces a principled cross-modal comparison: multi-source rollouts receive positive updates if and only if they outperform mono-source reasoning. This is consistent with the motivation that multi-source reasoning improves when it provides complementary information, while being regularized otherwise. Therefore, \modelname{} reduces the blind exploration of multi-source policies, accelerates convergence, and improves robustness against possible visual inconsistency, yielding a more stable and interpretable optimization trajectory for multi-source reasoning models as shown in Figure~\ref{fig:step}.

\noindent \textbf{Simplicity and Stability.}
Our algorithm only requires one policy model for optimization and does not introduce additional storage for models or samples, as opposed to experience sampling~\cite{zhan2026exgrpo} or off-policy correction~\cite{yan2025learning} methods. In practical implementation, we generate the mono-source samples by modifying image inputs and obtain the normalized statistics without computing gradients, thereby maintaining algorithmic efficiency and stability as shown in Table~\ref{tab:efficiency}.

\section{Experiments}
\label{sec:experiments}
\subsection{Experimental Setup}
\noindent \textbf{Datasets.}  Regarding the datasets, we employ diverse multi-source datasets. For visual modalities, in addition to typical RGB images, we incorporate four different modalities, including
depth SpatialQA~\cite{cai2025spatialbot}, infrared LLVIP~\cite{jia2021llvip}, multi-view nuScenes~\cite{bansal2020visual} and text-rich OCR-VQA~\cite{li2024llava}. Detailed explanations are shown in Appendix~\ref{app:dataset}.

\begin{table}[t]
\centering
\caption{\small Overall performance of four different multi-source tasks based on Qwen2.5-VL-3B. We compare with previous methods, supervised and reinforcement post-training methods, respectively. Union denotes the best performance among all mono-source results. The best results within a comparable group are in \textbf{Bold}.}
\vspace{3pt}
\label{tab:main-results}
\setlength{\tabcolsep}{2pt}  
\renewcommand{\arraystretch}{1.2} 
\resizebox{0.95\linewidth}{!}{
\begin{tabular}{lcccccccccc}
\toprule[1.3pt]
\multirow{2}{*}{\textbf{\ Model}} & \multicolumn{2}{c}{\textbf{Infrared}}&\multicolumn{2}{c}{\textbf{Depth}}&\multicolumn{2}{c}{\textbf{Multi-View}}&\multicolumn{2}{c}{\textbf{Text-Rich}}&\multicolumn{2}{c}{\textbf{Avg.}} \\
\cmidrule(lr){2-3} \cmidrule(lr){4-5} \cmidrule(lr){6-7} \cmidrule(lr){8-9} \cmidrule(lr){10-11}
 & \textbf{Union} &\textbf{Multi} & \textbf{Union} & \textbf{Multi} & \textbf{Union} & \textbf{Multi} & \textbf{Union} & \textbf{Multi} & \textbf{Union}& \textbf{Multi} \\
 \midrule
 \multicolumn{11}{c}{\textit{Previous Methods}} \\
 \midrule
Qwen2.5-VL-3B~\cite{bai2025qwen2} & 63.0 & 38.8 &86.3 & 73.8 & 82.0 & 74.0 & 69.0 & 51.0 & 75.1 & 59.4\\
LLaVA-OV-7B~\cite{li2024llavaonevision}& 4.9 & 2.1&80.0 &	56.3 & 88.0 & 44.0 & 81.0 & 67.0 &63.5 & 42.3 \\
R1-Onevision-7B~\cite{yang2025r1}& 19.7	& 11.1 & 87.5 & 72.5 & 90.0 & 81.0 & 77.0 & 58.0 & 68.5 & 55.6\\
VisionReasoner-7B~\cite{liu2025visionreasoner}& 82.5 & 75.2& 85.0 & 78.8 & 85.0 & 70.0 & 62.0 & 45.0 & 78.6 & 67.2\\
\midrule
\multicolumn{11}{c}{\textit{Supervised Post-Training}} \\
\midrule
SFT&87.1&84.5& 85.0 & 65.0 & 90.0 & 78.0 & 83.0 & 78.0 & 86.2 & 76.4\\
Chain-of-Thought& 88.2 & 84.2 & 90.0 & 77.5 & 90.0 & 75.0 & 77.0 & 76.0 & 86.3 & 78.2\\
\midrule
\multicolumn{11}{c}{\textit{Reinforcement Post-Training}} \\
\midrule
GRPO~\cite{shao2024deepseekmath} & 89.7 & 86.0 & 80.0 & 70.0 & 88.0	& 72.0	&77.0	&71.0& 83.6 & 74.7\\
\rowcolor{lightyellow} \textbf{GRPO+\modelname{}} & 89.9 & 87.7 & 80.0 & 73.8 & 90.0 & 75.0 & \textbf{79.0} & \textbf{75.0} & \textbf{84.7~\textcolor{Maroon}{(+1.1)}}& \textbf{77.9~\textcolor{Maroon}{(+3.2)}}\\
\midrule
DAPO~\cite{yu2025dapo}  & 88.8 & 85.5 & 76.3 & 73.8 & \textbf{92.0} & 74.0 &  75.0 & 70.0 & 83.0& 75.8\\
\rowcolor{lightyellow} \textbf{DAPO+\modelname{}} &\textbf{92.3} & \textbf{89.4} & \textbf{82.5} & \textbf{77.5} & 88.0 & \textbf{81.0} & 78.0 & \textbf{75.0} &\textbf{85.2~\textcolor{Maroon}{(+2.2)}}&\textbf{80.7~\textcolor{Maroon}{(+4.9)}}\\
\bottomrule[1.3pt]
\end{tabular}}
\vspace{-15pt}
\end{table}

\noindent \textbf{Baselines.}
For previous algorithms, we compare with various methods with~\cite{yang2025r1,liu2025visionreasoner} and without~\cite{li2024llavaonevision,bai2025qwen2} reinforcement post-training. In addition, for supervised post-training, SFT and CoT are also incorporated for comprehensive comparison. Regarding reinforcement post-training, we employ GRPO~\cite{shao2024deepseekmath} and DAPO~\cite{yu2025dapo}, which are two typical group-based reinforcement learning algorithms for visual reasoning. Since our method does not rely on a specific training framework, unless otherwise stated, all comparative experimental results are conducted within the same basic structure.

\noindent \textbf{Implementation details.}
We use Qwen2.5-VL-3B~\cite{bai2025qwen2} as the base model for supervised and reinforcement post-training. We mainly conduct the experiments on visual question answering~(VQA) and grounding, and the evaluation metrics are accuracy and mIoU, respectively. The concrete calculation is similar to that for verifiable rewards. For a comprehensive understanding, in addition to standard multi-source visual reasoning that jointly take all images as inputs~(\textbf{Multi}), we furthermore perform mono-source reasoning~(\textbf{Union}), \ie reason with each single source and then obtain the best result as final performance, as the upper bound to showcase the utility of information gain in performance. Concretely, we separately take every single source as input for visual reasoning during inference, and consider it to be correct if any single source correctly answers. We uniformly generate one trajectory for each visual source, \ie, $M=\texttt{image\_num}$ and $N=12$. More implementation details can be found in Appendix~\ref{app:detail}.

\subsection{Main Results}
We mainly conduct the experiment on multi-source datasets. Furthermore, we extend our algorithm to the application scenarios, the reinforcement post-training strategies and the scale of the model to comprehensively validate the effectiveness of the proposed method. More results are in Appendix~\ref{app:experiments}.

\noindent \textbf{\modelname{} is effective across various multi-source visual reasoning datasets.}
We evaluate the proposed advantage algorithm on diverse visual reasoning tasks with multi-source datasets. It can be seen in Table~\ref{tab:main-results} that: \textbf{(1)} the performance of Union exhibits better performance than Multi by a substantial margin across all tasks, which can be seen as the multi-source information gain, showing great potential to utilize for performance enhancement in multi-source tasks; \textbf{(2)} higher Union does not mean better performance in Multi, especially in supervised post-training methods, including SFT and CoT and naive RLVR methods. This strongly demonstrates the dynamic interaction between different sources is not well modeled by existing methods, leading to performance inferior to mono-source upper bound, which is consistent with our topic that \textit{seeing more does not mean knowing more in multi-source visual reasoning}; \textbf{(3)} compared to supervised post-training, better reinforcement learning strategy does not necessarily achieve better performance for Multi, revealing the difficulty of the task and highlighting the necessity for designed algorithms;
\textbf{(4)} in contrast to standard visual reasoning that even achieves no gains in some scenarios, our method consistently outperforms baselines and obtains substantial improvements on multi-source visual reasoning. Specifically, on the most important multi-source reasoning metric, Multi, our approach significantly boosts all tasks, with notable \textbf{3.8\%} and \textbf{7.0\%} improvements on infrared and multi-view datasets, respectively. Also, our method further facilitates the upper bound of Union, bringing in coherent improvements.
This confirms that dynamically leveraging mono-source rewards as anchors effectively guides the model to focus on informative sources while suppressing noise from less relevant images, and therefore achieves an impressive performance enhancement, firmly validating the effectiveness of our method.

\begin{figure}[t]
    \centering
    \includegraphics[width=1.0\linewidth]{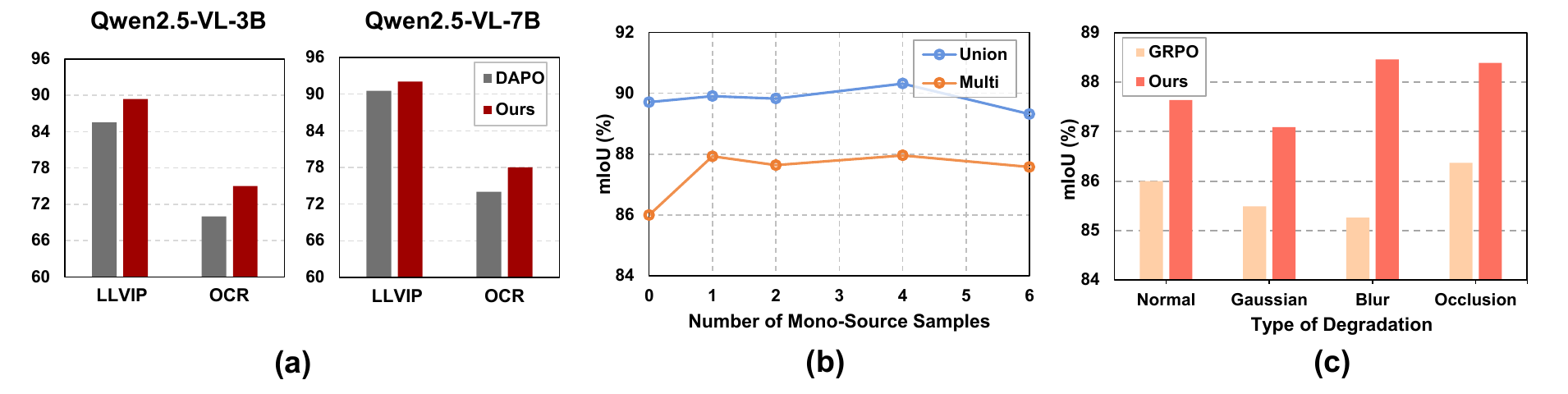}
    \vspace{-16pt}
    \caption{(a) Performance across different model sizes. (b) Influence of different numbers of mono-source samples in grounding. (c) Performance comparison under different visual degradations.}
    \vspace{-12pt}
    \label{fig:ablation}
\end{figure}

\noindent \textbf{Different training strategies.} Also, we validate the versatility of our approach by integrating it into two distinct reinforcement learning frameworks, \ie, GRPO and DAPO. As indicated in bottom of Table~\ref{tab:main-results}, our method obtains substantial improvements under both settings. Concretely, our method achieves an average performance gain of \textbf{3.2\%} in GRPO and \textbf{4.9\%} in DAPO on visual reasoning with various vision sources. Moreover, our algorithm also boosts the performance of mono-source reasoning by \textbf{1.1\%} and \textbf{2.2\%}, respectively. It highlights that our method is independent of the reinforcement learning framework, and only requires additional mono-source reward signals during advantage estimation without modifying the core training objectives, which shows great generalization ability and strong promise to serve as a plug-and-play integration to enhance multi-source performance in different training strategies.

\noindent \textbf{Generalizability across different model sizes.}
To further assess the scalability, we validate our method with model sizes comprising 3B and 7B parameters, respectively. As shown in Figure~\ref{fig:ablation}\textcolor{linkcolor}{a}, we can conclude that (1) larger models benefit from larger knowledge capacity and typically achieve a better performance; (2) the proposed advantage algorithm brings consistent and substantial improvement across scales, \eg, a 4.4\% gain for small model and  2.8\% for large model on average. This shows that the stable performance promotion stems from the inherent adaptability of the method by advantage estimation, making it agnostic to variation of model size, which suggests that the approach is not limited by model capacity and generalizes effectively. 

\subsection{Ablation Study and Further Analysis}
We conduct extensive ablation studies and in-depth analyses to validate the effectiveness of our method and provide insight into the contribution of the core components under different conditions.

\begin{figure}[h]
  \centering
  \vspace{-5pt}
  \begin{minipage}[t]{0.51\textwidth}
            \centering
            \captionsetup{type=table}
    \caption{Detailed statistics of reward in different algorithms.}
    \vspace{-3pt}
    \resizebox{\linewidth}{!}{\begin{tabular}{lcccccc}
    \toprule[1.3pt]
       \textbf{Statistics}  & \textbf{Qwen2.5-VL} &\textbf{GRPO} & \textbf{Ours}\\
       \midrule
       $\text{max}(r(o^\text{multi}))~(\uparrow)$&1.67 & 1.69 & \textbf{1.68} \\ 
       $\text{mean}(r(o^\text{multi}))~(\uparrow)$ &  1.49 & 1.51 & \textbf{1.62}\\ 
       $\text{max}(r(o^\text{mono}))~(\uparrow)$& 1.55 & 1.56 & \textbf{1.63} \\
    \bottomrule[1.3pt]
    \end{tabular}}
    \label{tab:statistics}
  \end{minipage}
  \hfill
  \begin{minipage}[t]{0.47\textwidth}
     \centering
     \captionsetup{type=table}
    \caption{\small performance and efficiency balance with different numbers of trajectory generation.}
    \renewcommand{\arraystretch}{1.25} 
    \resizebox{\linewidth}{!}{\begin{tabular}{cccc}
    \toprule[1.3pt]
       \textbf{Number of Trajectory}  & $N$ & $M+N$ & \textbf{Ours}\\
       \midrule
       \textbf{GPU Hours}   & 13$\times$8 & 17$\times$8 &14$\times$8\\
       \textbf{mIoU~(\%)} & 86.0 & 86.2 & 87.7\\
       \bottomrule[1.3pt]
    \end{tabular}}
    \vspace{-16pt}
    \label{tab:efficiency}
  \end{minipage}
\end{figure}

\noindent \textbf{Incorporating more mono-source samples.}
In the main results, we take one mono-source reward from every single source as the dynamic anchor. We investigate how the number of available mono-source samples, \ie, $M$, influences overall performance as the number of mono-source anchors employed in advantage normalization indicates the strength of the multi-source information gain towards robust visual reasoning. Results are shown in Figure~\ref{fig:ablation}\textcolor{linkcolor}{b}, where $M=0$ represents the vanilla algorithm. It reveals that visual reasoning accuracy saturates as $M$ increases and additional samples yield diminishing returns. 
It implies that a small set of mono-source trajectories is adequate to provide diverse information gain. Conversely, a larger $M$ does not necessarily bring gains, as mono-source samples may dominate the advantage estimation with unstable direction, leading to higher dynamics against multi-source reasoning in policy updates. Thus, we employ $M=2$, which is the number of sources, for stable training. More importantly,
guiding multi-source learning without requiring exhaustive sampling brings negligible computational or storage overhead, showing the efficiency.

\noindent \textbf{Impact of reward statistics.} We compare different statistics of reward in Table~\ref{tab:statistics} to investigate the changes in intuitive indicators. The results reveal that instead of increasing the best trajectory~(maximum multi-source reward), \modelname{} significantly improves the average quality of multi-source reasoning~(1.49 to 1.62), approaching the upper bound of mono-source reward. This shows that our method indeed exploits multi-source information gain to guide the optimization towards a better direction rather than stochastic exploration with diversity, which demonstrates the effectiveness of mono-source anchor in modeling the dynamic interaction. Also, the simultaneous improvement in mono-source trajectory~(1.55 to 1.63) during training uncovers the efficiency of the on-policy generation strategy without introducing additional model overload.

\noindent \textbf{Performance-efficiency trade-off.}
Our algorithm requires additional trajectory generation in practical implementation during reinforcement post-training. To certify the key component in effectiveness, we additionally perform GRPO with rollout to be $M+N$ and compare the training efficiency in Table~\ref{tab:efficiency}. The results show that our method, which includes only $N$ samples in advantage normalization, introduces moderate training overhead. By contrast, incorporating more actual trajectory in policy update brings marginal improvement~(0.2\%) with substantial overhead~(30\%).
This strongly suggests that the key to this approach is the mono-source anchor rather than introducing more rollouts, showcasing both effectiveness and efficiency.

\begin{figure}[t]
    \centering
    \includegraphics[width=1.0\linewidth]{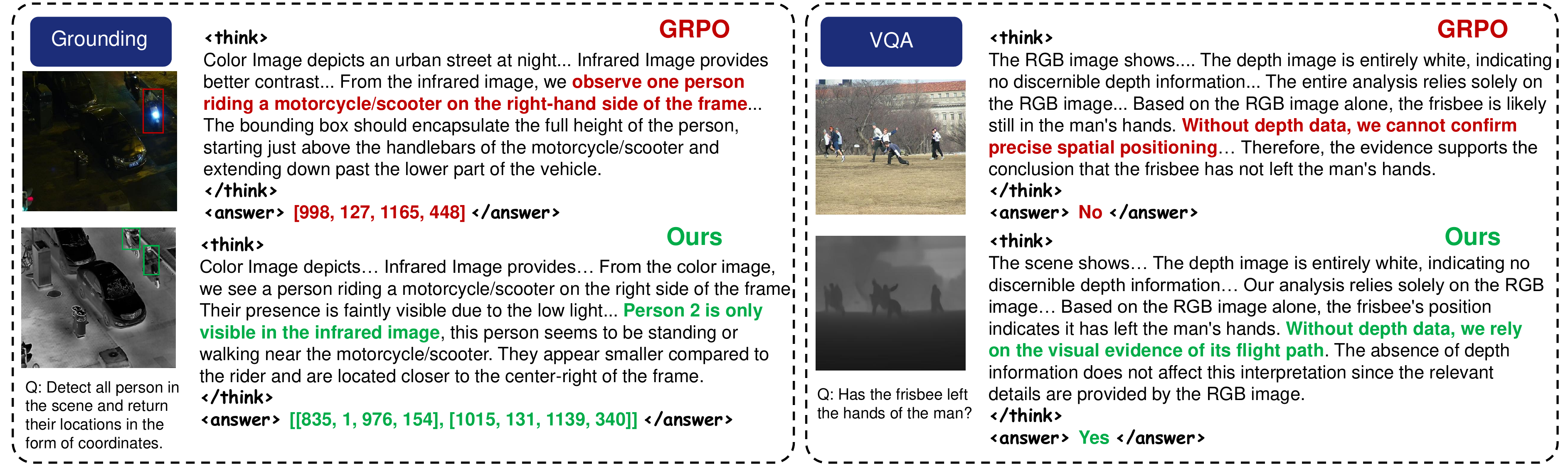}
    \caption{Qualitative results of visual reasoning on multi-source datasets. In grounding with RGB and infrared images and VQA with RGB and depth images, GRPO would excessively rely on the RGB image, resulting in improper predictions. Our method benefits from multi-source information gain and is capable of adaptively focusing on the key images with correct responses.}
    \vspace{-12pt}
    \label{fig:example}
\end{figure}

\noindent \textbf{Robustness to visual degradation.} As our method dynamically integrates interaction between modalities, and concentrates on a specific source if another displays no information gain, it shows great promise in exhibiting robustness to possible visual degradations of certain sources, \ie, noise, illumination, occlusion and so on. To simulate real-world conditions where visual inputs may be corrupted, we randomly degrade image quality by adding Gaussian noise, motion blur, and occlusion to the input images. As is illustrated in Figure~\ref{fig:ablation}\textcolor{linkcolor}{c}, our method maintains superior performance under all degradation types, \eg, only \textbf{0.5\%} drop in accuracy under severe Gaussian noise. Moreover, in the case of motion blur and occlusion, our method even obtains \textbf{0.8\%} performance gain, which outperforms the baseline by a substantial \textbf{1.3\%} improvement, showcasing its strong robustness and effectiveness. 
This can be attributed to our way of advantage normalization, which explicitly reduces the influence from unreliable source and relies more on stable, informative modality, thereby exhibiting adaptability and generalization ability.

\noindent \textbf{Case Study.} We visualize the reasoning process on several representative scenarios in Figure~\ref{fig:example} to qualitatively validate the effectiveness of our method. Specifically, in grounding task, our method successfully prioritizes key information in infrared image and detects the specific person that is hard to see in RGB, yielding more precise bounding boxes than GRPO. 
Moreover, in VQA task with RGB and depth images, while vanilla GRPO is uncertain about the answer without clearly analyzing depth image, our approach is confident about the inference relying solely on the RGB image and generates correct responses accordingly. The intuitive visualizations clearly show how our algorithm guides the model to weigh different sources appropriately, showcasing the effectiveness and superior capability.

\section{Conclusion}
In this paper, we revisit visual reasoning from a multi-source perspective, where visual modalities differing significantly in physical properties and semantics exhibit both promotion and conflict characteristics that are insufficiently modeled by existing methods. To address this issue, we propose a multi-source visual reasoning framework that adaptively emphasizes informative interaction while suppressing conflicting ones using mono-source rewards as dynamic anchors under RLVR training. This theoretically formulates the multi-source information gain of integration to guide the optimization towards stable and advanced trajectories. Comprehensive experiments across diverse datasets and training strategies with significant and consistent performance improvements strongly validate the effectiveness, efficiency and generalizability of the proposed approach. 

\bibliographystyle{plain}
\bibliography{neurips_2026}

\clearpage
\appendix

\section{Proofs}
\label{app:proof}
In this section, we provide a rigorous theoretical justification for the proposed method. We demonstrate that incorporating mono-source rewards as a dynamic baseline within the policy gradient framework yields an unbiased gradient estimator. Furthermore, we analyze how this mechanism effectively guides the optimization by adaptively weighting the objective based on multi-source information gain.

\subsection{Notations and Problem Formulation}
Consider a certain input with question $q$ and multiple images $i$. We define two distinct sets of rewards independently and identically distributed (\textit{i.i.d.}) from the same policy $\pi_\theta$:
(1) multi-source rollouts $\{o^{\text{multi}}_j\}_{j=1}^N$,
which follow standard reasoning procedure with joint images; and additionally (2) mono-source rollouts $\{o^{\text{mono}}_j\}_{j=1}^M$, which take only a subset of visual information,\ie, single-image to generate trajectories and subsequent rewards. Crucially, the reward depends solely on the input $q,i$ and is independent of the current trajectory $o$ generated by the policy.

Define the empirical means of the reward:
\begin{equation}
\mu^{\text{multi}} = \frac{1}{N}\sum_{j=1}^N r(q,i,o^{\text{multi}}_j),\ 
\mu^{\text{mono}}=\frac{1}{M}\sum_{j=1}^M r(q,i,o^{\text{mono}}_j).
\end{equation}
Moreover, the proposed algorithm constructs a dynamic baseline by concatenating mono-source and multi-source rollouts:
\begin{equation}
\mu^\text{hy}=\frac{M \mu^{\text{mono}} + N \mu^{\text{multi}}}{M+N}.
\end{equation}

The standard advantage function is obtained by performing normalization on the reward to reduce variance and regularize the training:
\begin{equation}
    A_j = \frac{r(q,i,o^{\text{multi}}_j) - \mu^{\text{multi}}}{\sigma^{\text{multi}}},
\end{equation}
where $\sigma$ is the variance of the reward for multi-source trajectories. In our algorithm, the advantage function is computed by replacing reward statistics with hybrid ones:
\begin{equation}
    A^\text{hy}_j = \frac{r(q,i,o^{\text{multi}}_j) - \mu^\text{hy}}{\sigma^\text{hy}}.
\end{equation}
The objective is to maximize the advantage function:
\begin{equation}
    \max_\theta J(\theta) = \mathbb{E}_{o\sim\pi_\theta}(A(o)).
\end{equation}
Here, we simplify the notation that the expectation of the advantage function performs on all samples and possible rollouts. Following the Policy Gradient Theorem \citep{sutton1998reinforcement}, the gradient estimation of the advantage function is:
\begin{equation}
    \nabla_\theta J(\theta) = \mathbb{E}_{\{q,i\}\sim  \mathcal{D}, \ o \sim \pi_\theta(q,i) } [A \cdot \nabla_\theta \log \pi_\theta(o|q,i)].
\end{equation}

Denote the multi-source information gain $\Delta_{IG} = \mu^\text{multi} - \mu^\text{mono}$ as the reward signal improvement brought by multiple visual modality input compared to mono-source input. $\Delta_{IG} > 0$ underlines positive benefits from multi-source visual reasoning and $\Delta_{IG} < 0$ suggests that multi-source reasoning introduces noise.

\begin{lemma}[Policy Gradient Identity]
\label{lem:score}
For any differentiable policy $\pi_\theta(\tau)$, we have:
\begin{equation}
\mathbb{E}_{o \sim \pi_\theta}
\left[
\nabla_\theta \log \pi_\theta(o)
\right]
= 0.
\end{equation}
\end{lemma}

\textit{Proof.} By definition,
\begin{equation}
\begin{aligned}
\mathbb{E}
\left[
\nabla_\theta \log \pi_\theta(\cdot)
\right] & =\int \pi_\theta(\tau) \nabla_\theta \log \pi_\theta(\tau)\, d\tau \\
&=
\int \pi_\theta(\tau)
\frac{\nabla_\theta \pi_\theta(\tau)}{\pi_\theta(\tau)}
\, d\tau \\
&=
\nabla_\theta
\int \pi_\theta(\tau)\, d\tau
=
\nabla_\theta 1
=
0.
\end{aligned}
\end{equation}
The first step expresses the objective in integral form over the distribution space. The second equation applies the \textit{log-derivative trick}, defined as $\nabla_\theta \log \pi_\theta(\cdot) = \frac{\nabla_\theta \pi_\theta(\cdot)}{\pi_\theta(\cdot)}$. The third equation follows the standard regularity conditions that for any differentiable policy, we can interchange the gradient and the integral operations. And finally, given by definition that $\pi_\theta$ is a valid probability distribution, its integral over the distribution space is always a constant.

\subsection{Proof Sketch of Theorem~\ref{the:unbiasedness}}
\label{app:proof1}
\begin{proof}
We proceed by expanding the expectation of policy gradient using the linearity of expectation:
\begin{equation}
\begin{aligned}
\nabla_\theta &J^\text{hy}(\theta) = \mathbb{E}_{q,i\sim  \mathcal{D}, \ o \sim \pi_\theta(q,i) }[{A^\text{hy}}\cdot \nabla_\theta\log \pi_\theta(o|q,i)] \\
&= \mathbb{E}_{q,i\sim  \mathcal{D}, \ o \sim \pi_\theta(q,i) }[r(q,i,o^{\text{multi}})\cdot \nabla_\theta\log \pi_\theta(o|q,i)] \\
& \qquad \qquad -\mathbb{E}_{q,i\sim  \mathcal{D}, \ o \sim \pi_\theta(q,i) }[\mu^\text{hy}\cdot \nabla_\theta\log \pi_\theta(o|q,i)].
\end{aligned}
\end{equation}
By the definition of gradient estimation and Lemma.~\ref{lem:score}, the first term corresponds exactly to the policy gradient of the standard advantage function $\nabla_\theta J(\theta)$ and the second term equals zero~\cite{williams1992simple}. Thus, the expectation of the gradient with a dynamic mono-source anchor remains equal to the original gradient, which completes the proof.
\end{proof}
We prove that for any baseline function that depends on the input sampled from the current policy $\pi_\theta$ does not introduce bias to gradient estimation, and also holds for the proposed advantage normalization in our algorithm. This provides a theoretical guarantee for the stability of our algorithm.

\subsection{Proof Sketch of Theorem~\ref{the:decom}}
\label{app:proof2}
\begin{proof}
The standard gradient estimation~(ignoring constant scaling $\sigma$) can be formulated as:
\begin{equation}
    \nabla_\theta J(\theta) = \frac{1}{N}\sum_{j=1}^N(r(q,i,o^\text{multi}_j)-\mu^\text{multi})\nabla_\theta \log \pi_\theta(o^\text{multi}_j|q,i),
\end{equation}
where standard advantages have a zero mean:
\begin{equation}
\sum_{j=1}^{N} (r(q,i,o^\text{multi}_j)-\mu^\text{multi}) = 0,
\end{equation}
that guarantees reduced gradient variance. Consider the gradient under hybrid advantage normalization:
\begin{equation}
    \nabla_\theta J^\text{hy}(\theta) = \frac{1}{N}\sum_{j=1}^N(r(q,i,o^\text{multi}_j)-\mu^\text{hy})\nabla_\theta \log \pi_\theta(o_j^\text{multi}|q,i).
\end{equation}

In the sense of expectation, the mean of the proposed algorithm can be regarded as a mixture distribution over mono-source and multi-source rollouts:
\begin{equation}
    \mu^\text{hy} = \alpha \mu^\text{multi} + (1-\alpha)\mu^\text{mono},
\end{equation}
where $\alpha=N/(M+N)$ is the proportion of multi-source trajectories. Therefore, expanding the normalization term:
\begin{equation}
\begin{aligned}
r(&o^\text{multi}_j)-\mu^\text{multi} \\
= &\ r(o^\text{multi}_j)-\big(\alpha \mu^\text{multi} + (1-\alpha)\mu^\text{mono}\big)\\
= &r(o^\text{multi}_j)-\big(\mu^\text{multi}  - (1-\alpha) \mu^\text{multi} + (1-\alpha)\mu^\text{mono}\big)\\
= &(r(o^\text{multi}_j)-\mu^\text{multi})  + (1-\alpha) (\mu^\text{multi} -\mu^\text{mono})\\
=&A_j  + (1-\alpha)\Delta_{IG}.\\
\end{aligned}
\end{equation}

Substituting the simplified term into the original expression, the hybrid gradient can be decomposed into two parts:
\begin{equation}
\begin{aligned}
\nabla_\theta J^\text{hy}(\theta)
 = &\underbrace{\frac{1}{N}\sum_{j=1}^NA_j\nabla_\theta \log \pi_\theta(o^\text{multi}_j|q,i)}_{\text{Original GRPO gradient}}\\
 +(1-&\alpha)\underbrace{\Delta_{IG}\Big(\frac{1}{N}\sum_{j=1}^N\nabla_\theta \log \pi_\theta(o^\text{multi}_j|q,i)\Big)}_{\text{Information gain regularization}}.
\end{aligned}
\label{eqn:generalization-bound}
\end{equation}
It can be concluded that the gradient estimation of our algorithm is to optimize the standard gradient weighted with information gain regularization during training, where the first term refers to the original GRPO gradient estimation, and the second term is a regularization term in proportion to the relative information gain. This completes the proof.
\end{proof}

We reveal that the method amplifies the relative performance signal between multi-source and mono-source reasoning by anchoring a mono-source baseline, preserves unbiased policy gradients, and inherits standard convergence guarantees. Concretely, (1) if the mono-source baseline performs well, $\mu^\text{hy}$ increases. This sets a higher bar for the gradient policy. If the multi-source model generates a response with a reward lower than the mono-source one, the advantage $A^\text{hy}$ becomes negative. Consequently, the optimization suppresses the probability of this trajectory; (2) if the mono-source anchor is low, $\mu^\text{hy}$ decreases. The model is effectively rewarded for any marginal improvement over mono-source performance, encouraging effective utilization of multi-source cues.

\section{More Details of Experiments}
\subsection{Implementation details.}
\label{app:detail}
We build the codebase for visual reasoning based on verl~\footnote{https://github.com/volcengine/verl}. The main experiments are trained on NVIDIA A800 for 2 epochs to get a fair comparison. In evaluation, we separately reason with every single source and get the best result for each question as the final mono-source performance to see the effectiveness of the mono-source anchor. The results are constructed by extracting the prediction from the answer part \texttt{<answer>$\cdots$<\/answer>}. 

\subsection{Datasets.}
\label{app:dataset}
\noindent \textbf{LLVIP} is an object detection dataset. It is composed of RGB and infrared images corresponding to the same scene. It contains a large number of scenes with poor light conditions at night with rich pedestrian annotations, validating the robustness of cross-modal perception under challenging illumination scenarios. We therefore use it as the RGB \& Infrared dataset.

\noindent \textbf{SpatialQA} is a robotic manipulation dataset that aims to precisely comprehend spatial information through depth images. It is composed of corresponding RGB-Depth image pairs with multi-level questions across various embodiment tasks and scenarios, and we exploit it to evaluate RGB \& Depth polar questions.

\noindent \textbf{nuScenes} is a multimodal VQA dataset for autonomous driving scenarios. For each sample in the driving scenarios, it consists of six surrounding images and question-answer pairs based on visual scenes and question templates. Its different perspectives can be seen as different sources of information, so we employ it to evaluate the performance of visual reasoning models in multi-view scenarios.

\noindent \textbf{OCR-VQA} is a visual question answering dataset focusing on scene text understanding. It contains real-world images with rich embedded text, which requires accurate reasoning over the recognized text. We use the dataset to evaluate text-centric visual reasoning with multiple-choice questions.

\section{More Experimental Results}
\label{app:experiments}
\noindent \textbf{Results on general multimodal benchmarks.}
We also comprehensively evaluate the general capabilities of different methods including single-image benchmarks MMstar~\cite{chen2024we}, HalluBench~\cite{guan2024hallusionbench} TextVQA~\cite{singh2019towards} and multi-image benchmarks Mantis~\cite{jiang2024mantis}, Muirbench~\cite{wang2025muirbench} Blink~\cite{fu2024blink}. It can be seen from Table~\ref{tab:general} that our method also preserve the capability in general multi-modal benchmarks with average \textbf{1.7\%} improvement, showing the generalizability.

\begin{table}[h]
    \centering
    \vspace{-10pt}
    \caption{Comparison on general multimodal benchmarks.}
    \resizebox{0.96\linewidth}{!}{
    \begin{tabular}{lccccccc}
    \toprule[1.3pt]
        \multirow{2}{*}{\textbf{Method}} &\multicolumn{3}{c}{\textbf{Single-Image}} &\multicolumn{3}{c}{\textbf{Multi-Image}}& \multirow{2}{*}{\textbf{Avg.}}\\
        \cmidrule(lr){2-4} \cmidrule(lr){5-7}
         & \textbf{MMStar} & \textbf{HalluBench} & \textbf{TextVQA} & \textbf{Mantis} & \textbf{Muirbench} & \textbf{Blink}  & \\
        \midrule
        Qwen2.5-VL-3B &54.5& \textbf{60.6}&\textbf{79.1}& 59.4&\textbf{49.0}&53.8&59.4 \\
        GRPO & 55.2&57.6&75.5&57.1&47.3&81.7&62.4\\
        \textbf{\modelname{}} & \textbf{55.9}&58.4&76.0&\textbf{62.7}&48.9&\textbf{82.5}&\textbf{64.1~\textcolor{Maroon}{(+1.7)}}\\
        \bottomrule[1.3pt]
    \end{tabular}}
    \vspace{-5pt}
    \label{tab:general}
\end{table}

\noindent \textbf{More analysis of computational overhead.}
In our method, mono-source trajectory is generated for each visual source. Therefore, the efficiency when incorporating more visual sources should be carefully discussed. To better illustrate this, we increase the number of mono-source samples $M$ in nuScenes, where mono-source rollouts are generated for six visual sources in training sample and report wall-clock time and GPU memory overhead to analyze its impact on computational cost.

The results in Table~\ref{tab:additional-rollout} show that adding more mono-source samples brings performance saturation~(75.0$\rightarrow$76.0), negligible increase in GPU memory~(52 GB$\rightarrow$54 GB) and training time~(11.1$\times$8 h$\rightarrow$11.5$\times$8 h) without additional GPU memory consumption in backpropagation~(80 GB). This indicates that mono-source samples are not computational bottleneck, which can be attributed to the anchor function of mono-source rewards that are not actually engaged in loss backpropagation. The major computational cost is still backward-pass of $N$ multi-source rewards, strongly validating the effectiveness and efficiency of the proposed method.

\begin{table}[h]
    \vspace{-10pt}
    \centering
    \begin{minipage}{0.59\linewidth}
        \caption{Detailed analysis of computational overhead.}
    \resizebox{\linewidth}{!}{
    \begin{tabular}{lccc}
        \toprule[1.3pt]
        $M$&\textbf{0}&\textbf{6}&\textbf{12}\\
        \midrule
         Performance & 72.0&75.0&76.0\\
         GPU Memory~(rollout generation)& 50 GB&52 GB&54 GB \\
         GPU Memory~(backpropagation)&80 GB&80 GB&80 GB \\
         Wall-Clock Training Hours &10.6$\times$8 & 11.1$\times$8 &11.5$\times$8\\
         \bottomrule[1.3pt]
    \end{tabular}}
    \label{tab:additional-rollout}
    \end{minipage}
    \begin{minipage}{0.39\linewidth}
        \centering
    \caption{Influence of different anchors.}
    \resizebox{\linewidth}{!}{
    \begin{tabular}{lcccc}
        \toprule[1.3pt]
        \multirow{2}{*}{\textbf{Method}} & \multicolumn{2}{c}{\textbf{GRPO}} & \multicolumn{2}{c}{\textbf{DAPO}}\\
        \cmidrule(lr){2-3} \cmidrule(lr){4-5} 
        &\textbf{Union} & \textbf{Multi}&\textbf{Union} & \textbf{Multi}\\ 
        \midrule
        Fixed & 88.1 & 84.8 &88.9 & 85.2\\
        Weighted & 86.9 & 79.7& 86.2 & 80.6\\
        \textbf{Ours}& 89.9& 87.7 &92.3 & 89.4 \\
        \bottomrule[1.3pt]
    \end{tabular}}
    \label{tab:anchor}
    \end{minipage}
    \vspace{-5pt}
\end{table}

\noindent \textbf{Comparison to alternative anchors.}
To validate the effectiveness of the proposed anchor, we replace our dynamic mono-source anchor with a fixed value function or a weighted average reward of different modalities. In Table~\ref{tab:anchor}, both alternatives lead to a significant performance drop from 2\% to 10\%. By contrast, our approach that integrates concurrent single- and multi-source rewards with simple average theoretically ensures consistent and context-aware advantage estimation, underscoring the importance of using an on-policy, rollout-based anchor.

\noindent \textbf{Per-source results.} Figuring out which source is strong and the relation with multi-source reasoning is important for understanding the effectiveness. Therefore, we conduct experiments on infrared and depth and report the performance of each source and to show the influence of different sources. Results are shown in Table~\ref{tab:per-source}. It can be concluded that (1) infrared and depth images can provide reliable evidence for certain scenes, so they are strong and can obtain better performance than RGB, which is consistent with the motivation of the paper; (2) our method consistently captures reliable modality, achieves better performance in each modality, and as a result gets better performance in Multi, showing the effectiveness.

\begin{table}[h]
    \vspace{-10pt}
    \centering
    \caption{Per-source results on infrared and depth.}
    \resizebox{0.95\linewidth}{!}{
    \begin{tabular}{lcccccccc}
        \toprule[1.3pt]
        \multirow{2}{*}{\textbf{Method}}&\multicolumn{4}{c}{\textbf{Infrared}} &  \multicolumn{4}{c}{\textbf{Depth}}\\
        \cmidrule(lr){2-5}  \cmidrule(lr){6-9}
         & \textbf{RGB-only} & \textbf{Infrared-only} & \textbf{Multi} & \textbf{Union} & \textbf{RGB-only} & \textbf{Depth-only} & \textbf{Multi} & \textbf{Union}\\ 	
         \midrule
        Qwen2.5-VL-3B & 46.7 & 51.8 & 38.8 & 63.0 & 65.0 & 52.7 & 73.8 & 86.3\\
        DAPO & 71.9 & 82.0 & 85.5 & 88.8 & 62.4 & 68.7 & 73.8 & 76.3\\ 
        DAPO+\modelname{} & 73.1 & 83.5 & 89.4 & 92.3 & 65.1 & 71.2 & 77.5 & 82.5 \\
        \bottomrule[1.3pt]
    \end{tabular}}
    \vspace{-5pt}
    \label{tab:per-source}
\end{table}

\begin{figure}[t]
    \centering
    \includegraphics[width=1.0\linewidth]{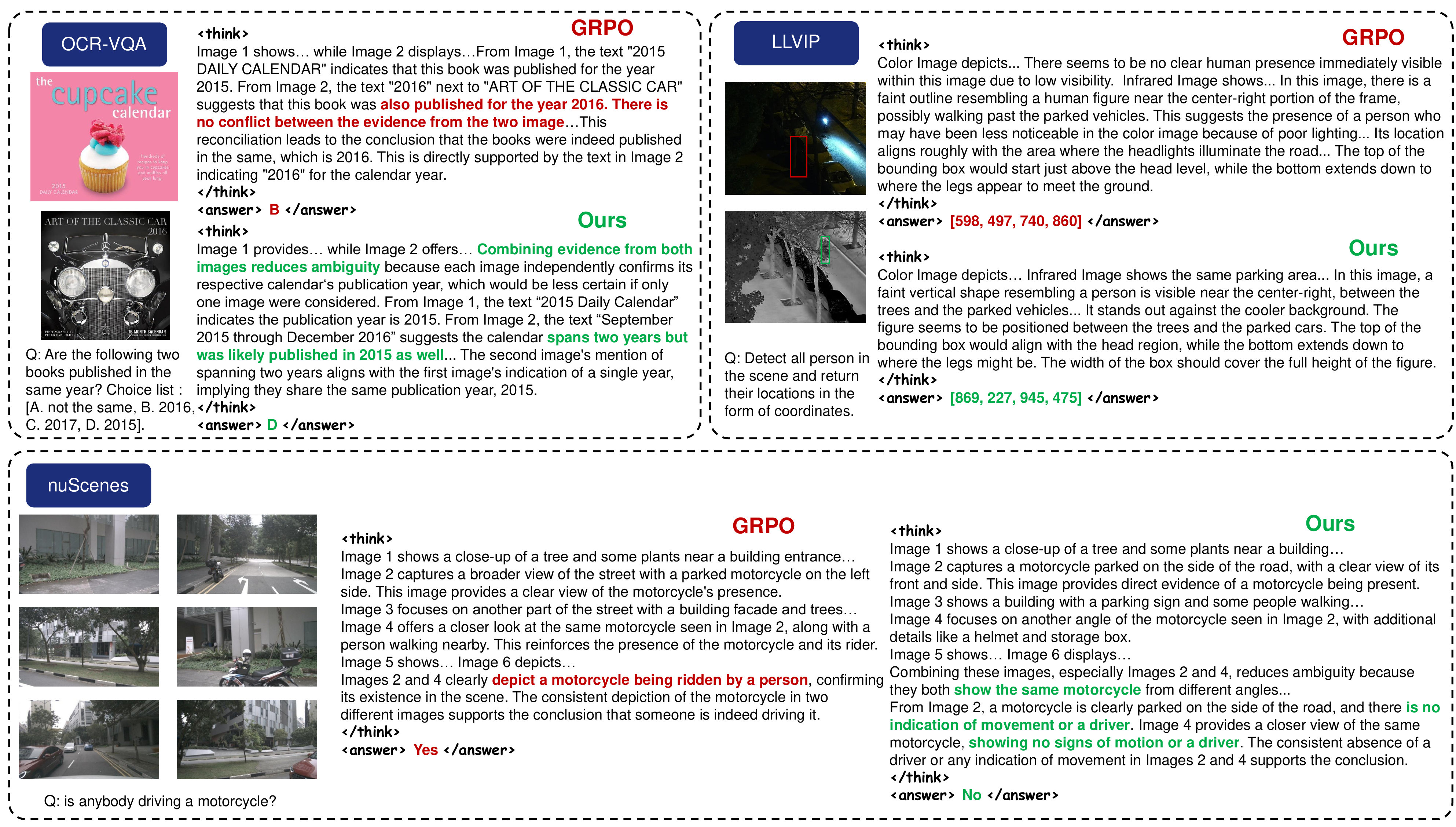}
    \caption{More qualitative results of visual reasoning on OCR-VQA, LLVIP and nuScenes.}
    \vspace{-5pt}
    \label{fig:more-example}
\end{figure}

\noindent \textbf{More qualitative results.} We visualize more comparison examples in Figure~\ref{fig:more-example} to show that our method effectively focuses on specific source and gives proper responses accordingly in different kinds of occasions, which further validates the effectiveness of our method.

\section{Limitations and Future Directions}
This work proposes a multi-source visual reasoning framework for MLLMs that explicitly models and quantifies multi-source information gain from modality integration and effectively improves the performance. Despite the promise, several issues are to be addressed. One primary limitation is that due to resource constraints, our evaluation is mainly conducted on moderately sized multimodal models and does not extend to more types of foundation models and reinforcement learning algorithms. Nevertheless, since our approach is general and architecture-agnostic, we believe it can generalize well. Moreover, our study focuses on different visual modalities and does not directly address broader multimodal scenarios. Real-world reasoning tasks would also involve additional modalities such as video, audio, and 3D point clouds, which introduce temporal and structural complexity. We believe the core principle of dynamic anchor can generalize beyond multi-source visual reasoning and would treat adapting our framework to these richer multimodal settings as a future direction.

\end{document}